\documentclass[openacc]{rstransa}%%%%where rstrans is the template name

\usepackage{amsmath, amssymb, bm, mathtools}
\usepackage{microtype}

\usepackage[round, authoryear]{natbib}
\usepackage{hyperref}
\hypersetup{colorlinks=false, hidelinks}

\usepackage[lined,linesnumbered,ruled,vlined,algo2e]{algorithm2e}
\SetKw{KwForIn}{in}
\SetKw{KwUntil}{until}
\SetKwInput{KwInputs}{Inputs}
\SetKwInput{KwOutputs}{Outputs}

\newcommand{\cu}[1]{
	\ifcat\noexpand#1\relax
	\bm{#1}
	\else
	\mathbf{#1}
	\fi
}

\newcommand{\diff}{\mathop{}\!\mathrm{d}}

\newcommand{\cond}{{\;|\;}}

\let\lim\relax
\DeclareMathOperator*{\argmin}{arg\,min\,}  % Argmin
\DeclareMathOperator*{\lim}{lim\,}  % inf
\let\grad\relax
\DeclareMathOperator{\grad}{\nabla\!}

\newcommand{\expecsym}{\operatorname{\mathbb{E}}}     % Expec
\newcommand{\covsym}{\operatorname{Cov}}     % Covariance
\newcommand{\varrsym}{\operatorname{Var}}     % Variance
\newcommand{\diagsym}{\operatorname{diag}}     % Diagonal matrix
\newcommand{\tracesym}{\operatorname{tr}}           % Trace

\let\expec\relax
\let\cov\relax
\let\varr\relax
\let\diag\relax
\let\trace\relax

\makeatletter
\newcommand{\expec}{\@ifstar{\@expecauto}{\@expecnoauto}}
\newcommand{\@expecauto}[1]{\expecsym \left[ #1 \right]}
\newcommand{\@expecnoauto}[1]{\expecsym [#1]}

\newcommand{\expecbigg}[1]{\expecsym \biggl[ #1 \biggr]}

\newcommand{\cov}{\@ifstar{\@covauto}{\@covnoauto}}
\newcommand{\@covauto}[1]{\covsym \left[ #1 \right]}
\newcommand{\@covnoauto}[1]{\covsym [#1]}

\newcommand{\varr}{\@ifstar{\@varrauto}{\@varrnoauto}}
\newcommand{\@varrauto}[1]{\varrsym \left[ #1 \right]}
\newcommand{\@varrnoauto}[1]{\varrsym [#1]}

\newcommand{\diag}{\@ifstar{\@diagauto}{\@diagnoauto}}
\newcommand{\@diagauto}[1]{\diagsym \left( #1 \right)}
\newcommand{\@diagnoauto}[1]{\diagsym (#1)}

\newcommand{\trace}{\@ifstar{\@traceauto}{\@tracenoauto}}
\newcommand{\@traceauto}[1]{\tracesym \left( #1 \right)}
\newcommand{\@tracenoauto}[1]{\tracesym (#1)}

\makeatother

\newcommand*{\trans}{{\mkern-1.5mu\mathsf{T}}}

\newcommand*{\R}{\mathbb{R}} % Set of real numbers
\newcommand*{\Q}{\mathbb{Q}} % Set of rational numbers
\newcommand*{\bbP}{\mathbb{P}}

\newcommand{\normbig}[1]{\bigl\lVert #1 \bigr\rVert}

\let\kl\relax
\newcommand{\kl}[2]{\mathrm{KL}( #1 \, \Vert \, #2 )}
\newcommand{\klbig}[2]{\mathrm{KL}\bigl( #1 \, \Vert \, #2 \bigr)}

\makeatletter
\@ifundefined{thmenvcounter}{}
{%
	\newtheorem{envcounter}{EnvcounterDummy}[\thmenvcounter]

	\newtheorem{remark}[envcounter]{Remark}

	\newtheorem{algorithm}[envcounter]{Algorithm}
	
}
\makeatother
\newcommand{\refmeasure}{\pi_{\mathrm{ref}}}

\newtheorem{algorithm}{\bf Algorithm}[section]

\titlehead{Review}

\begin{document}

%%%% Article title to be placed here
\title{Conditional sampling within generative diffusion models}

\author{%%%% Author details
Zheng Zhao$^{1,2}$, Ziwei Luo$^2$, Jens Sj\"{o}lund$^2$, and Thomas B. Sch\"{o}n$^2$}

%%%%%%%%% Insert author address here
\address{$^1$Department of Computer and Information Science, Link\"{o}ping University, Sweden\\$^2$Department of Information Technology, \\Uppsala University, Sweden}

%%%% Subject entries to be placed here %%%%
\subject{Computational statistics}

%%%% Keyword entries to be placed here %%%%
\keywords{Generative diffusions, stochastic differential equations, conditional sampling, Bayesian inference}

%%%% Insert corresponding author and its email address}
\corres{Zheng Zhao\\
\email{zheng.zhao@liu.se}}

%%%% Abstract text to be placed here %%%%%%%%%%%%
\begin{abstract}
Generative diffusions are a powerful class of Monte Carlo samplers that leverage bridging Markov processes to approximate complex, high-dimensional distributions, such as those found in image processing and language models. 
Despite their success in these domains, an important open challenge remains: extending these techniques to sample from conditional distributions, as required in, for example, Bayesian inverse problems. 
In this paper, we present a comprehensive review of existing computational approaches to  conditional sampling within generative diffusion models. 
Specifically, we highlight key methodologies that either utilise the joint distribution, or rely on (pre-trained) marginal distributions with explicit likelihoods, to construct conditional generative samplers.

This article is part of the theme issue ``Bayesian inverse problems with generative models''.
\end{abstract}
%%%%%%%%%%%%%%%%%%%%%%%%%%%

%%%%%%%%%% Insert the texts which can accomdate on firstpage in the tag "fmtext" %%%%%

\begin{fmtext}

% empty

\end{fmtext}

%%%%%%%%%%%%%%% End of first page %%%%%%%%%%%%%%%%%%%%%

\maketitle

\section{Introduction}
\label{sec:intro}

Consider the conditional probability distribution $\pi(\cdot \cond y)$ of a random variable $X\in\R^{d}$ with condition $y\in\R^{d_y}$. 
Sampling the distribution is the fundamental question in computational statistics, and there are a plethora of developed sampling schemes to use depending on what we know about  $\pi(\cdot \cond y)$. 
As an example, when the density function of $\pi(\cdot \cond y)$ is available (up to a constant), Markov chain Monte Carlo~\citep[MCMC,][]{Meyn2009} methods are popular and generic algorithms widely used. 
The MCMC algorithms simulate a Markov chain that is invariant in the target distribution. The drawback is that this often makes the algorithms computationally and statistically inefficient for high-dimensional problems. 

In this article, we discuss an emerging class of samplers that leverage \emph{generative diffusions}~\citep[see, e.g.,][]{Benton2024, Song2021scorebased, Daras2024survey}, which have empirically worked well for many Bayesian inverse problems.  
At the heart, the generative diffusions aim to find a continuos-time Markov process (e.g., stochastic differential equation) that bridges the target distribution and a reference distribution, so that sampling the target simplifies to sampling the reference and the Markov process. 
In contrast to traditional samplers such as MCMC which use the target's density function to build statistically exact samplers, the generative diffusions use the data to approximate a sampler akin to normalising flow~\citep{Chen2018, Papamakarios2021} and flow matching~\citep{Lipman2023flow}. 
This comes with at least three benefits compared to MCMC: 1) scalability of the problem dimension (after the training time), 2) no need to explicitly know the target density function, 3) and the sampler is differentiable~\citep[see, e.g., a use case in][]{Watson2022learning}.

However, the generative diffusion framework (for unconditional sampling) is not immediately applicable to conditional sampling, since we do not have the conditional data samples from $\pi(\cdot \cond y)$ required to train the generative samplers. 
This article thus presents a class of generative samplers that utilise data samples from the joint $\pi_{X,Y}$, or the marginal $\pi_X$ with explicit likelihood, which are typically more accessible than the conditional samples. 
Following this class of constructions, plenty of generative conditional samplers have been developed specifically in their respective applications, for instance, the guided diffusions in computer vision~\citep{Luo2024DACLIP, Kawar2022, Lugmayr2022CVPR}, where they can sample images with additional guidance inputs (e.g., text instructions).
In this article, we focus on describing generic methodologies and show how they can inspire new ones. 

The article is organised as follows. 
In the following section, we briefly explain the core of generative diffusions to set-up the preliminaries of generative conditional samplers. 
Then, in Section~\ref{sec:cond-sampling-joint} we show how to approximate the samplers based on the data sampled from the joint distribution, called the joint bridging methods. 
In Section~\ref{sec:cond-sampling-likelihood} we show another important approach using Feynman--Kac models that leverages the data from the marginal when the likelihood model is accessible, followed by a pedagogical example.
%The target audience are statisticians who are familiar with SDEs but not much on generative SDEs.

\paragraph{Notation} If $\lbrace X(t) \rbrace_{t=0}^T$ is a forward process, then we abbreviate $p_{t \cond s}(\cdot \cond x)$ as the distribution of $X(t)$ conditioned on $X(s) = x$ for $t>s$. 
If the process is further correlated with another random variable $Y$, then we use $p_{t, Y}$ as the joint distribution of $(X(t), Y)$, and $p_{t \cond Y}(\cdot \cond y)$ as the conditional one. 
We apply this notation analogously for the reverse process $\lbrace U(t) \rbrace_{t=0}^T$, but use $q$ instead of $p$, e.g. $q_{t \cond s}(\cdot \cond u)$ for $U(t)$ conditioned on $U(s) = u$. 
The path measures of $\lbrace X(t) \rbrace_{t=0}^T$ are denoted by $\bbP$ with interchangeable marginal notation $\bbP_t \equiv p_t$. 

\section{Generative diffusion sampling}
\label{sec:generative-sampling}
We begin by explaining how generative diffusions are applied to solve unconditional sampling. 
Let $\pi$ be the distribution of a random variable $X\in\R^d$ that we aim to sample from. 
The generative diffusions start with a (forward-time) stochastic differential equation (SDE)
\begin{equation}
	\diff X(t) = a(X(t), t) \diff t + b(t) \diff W(t), \quad X(0) \sim \pi, \quad X(T)\sim p_T, 
	\label{equ:fwd}
\end{equation}
that continuously transports $\pi$ to another reference distribution $p_T$ at time $T$, where $a\colon\R^d\times[0,T]\to \R^d$ and $b\colon[0,T]\times\R^{d\times w}$ are (unknown) drift and dispersion functions, and $W\in\R^w$ is a Brownian motion. 
Then, the gist is to find a reverse correspondence of Equation~\eqref{equ:fwd} such that if the reversal initialises at distribution $p_T$ then it ends up with $\pi$ at $T$. 
Namely, we look for a reversal
\begin{equation}
	\diff U(t) = f(U(t), t) \diff t + g(t) \diff W(t), \quad U(0) \sim p_T, \quad U(T) \sim \pi,
	\label{equ:bwd}
\end{equation}
and from now on we use $q_t$ to denote the marginal law of $U(t)$. 
If the forward SDE coefficients in Equation~\eqref{equ:fwd} are designed in such a way that the implied $p_T$ is easy to sample (e.g., a Gaussian), then we can sample the target $\pi$---which is hard---by simulating the reversal in Equation~\eqref{equ:bwd} which is simpler.\footnote{The Brownian motions in Equations~\eqref{equ:fwd} and~\eqref{equ:bwd} are not necessarily the same. 
To avoid clutter, we throughout the paper apply the same notation $W$ for all Brownian motions when we are not concerned with distinguishing them.} 

There are infinitely many such reversals if we only require that the forward and reversal match their marginals at $t=0$ and $t=T$. 
However, if we also constrain the path of the marginal distribution $t\mapsto p_t$ to be equal to $t\mapsto q_{T-t}$, then we arrive at the classical result by~\citet{Anderson1982} allowing us to explicitly write the reversal as 
\begin{equation}
	f(u, t) = -a(u, T - t) + \Gamma(T - t)\grad\log p_{T-t}(u), \quad g(t) = b(T - t),
	\label{equ:rev-drift-anderson}
\end{equation}
where we shorten $\Gamma(t) \coloneqq b(t) \, b(t)^\trans$, and $p_{t}$ here stands for marginal law of $X(t)$. 
In other words, Anderson's reversal here means that it solves the same Kolmogorov forward equation (in reverse time) as with Equation~\eqref{equ:fwd}. 

However, it is known that Anderson's construction comes with two computational challenges. 
The first challenge lies in the intractability of the score function $\grad\log p_t$ (since the score of $\pi$ is unknown).  
To handle this, the community has developed means of numerical techniques to approximate the score. 
A notable example in this context is denoising score matching~\citep{Song2021scorebased}. 
It works by parametrising the score $\grad\log p_t(x) = r(x, t; \eta)$ with a neural network, and then estimate the parameter $\eta$ by solving an optimisation problem
\begin{equation}
	\argmin_{\eta\in\R^{d_\eta}} \expecbigg{\int_0^T \normbig{r(X(t), t; \eta) - \grad_{X(t)}\log p_{t\cond 0}(X(t) \cond X(0))}_2^2 \diff t}, 
	\label{equ:score-matching}
\end{equation}
where the expectation takes on the measure of the forward SDE in Equation~\eqref{equ:fwd}. 
Note that the forward transition $p_{t\cond 0}$ is fully tractable when we choose $a$ as a linear function, and a convergence analysis is provided by~\citet[][Thm. 1]{DeBortoli2021diffusion}. 
The second challenge is that it is non-trivial to design a forward SDE that \emph{exactly} terminates at an easy-to-sample $p_T$. 
In practice, we often specify~$p_T$ as a Gaussian and then design the forward equation as the associated Langevin dynamics, at the cost of \emph{assuming} large enough $T$. 
Another challenge consists in the discretisation errors when numerically solving Equation~\eqref{equ:bwd}, requiring finer time grid for more accurate simulation. 
Significant work has gone into resolving this computational  problem, such as distillation~\citep{Salimans2022progressive}.

The so-called dynamic Schr\"{o}dinger bridge (DSB) is gaining traction as an alternative to Anderson's construction~\citep{DeBortoli2021diffusion}. 
In this way, Equation~\eqref{equ:fwd} can be obtained as the solution to a DSB with a fixed $p_T$. 
Suppose that we aim to bridge between $\pi$ and any specified reference measure $\pi_{\mathrm{ref}}$ (here refers to $p_T$), the DSB constructs Equation~\eqref{equ:fwd} (and its reversal) as the unique solution
\begin{equation}
	\bbP^\star = \argmin_{\substack{\bbP_0 = \pi, \bbP_T=\pi_{\mathrm{ref}} \\ \bbP \in \mathcal{M} }}\kl{\bbP}{\mathbb{Q}},
	\label{equ:dsb}
\end{equation}
where $\mathbb{Q}$ is the path measure of a reference process (not related to the reference $\pi_{\mathrm{ref}}$), and $\mathcal{M}$ stands for the collection of Markov path measures that are solutions to SDEs of the form in Equation~\eqref{equ:fwd}. 
Namely, DSB searches for diffusion processes that \emph{exactly} bridge $\pi$ and the given $\pi_{\mathrm{ref}}$, and then DSB chooses the unique one that is closest to the reference process in the sense of Kullback--Leibler divergence. 
The solution of Equation~\eqref{equ:dsb} is typically not available in closed form, and it has to be computed numerically. 
Compared to Anderson's construction, DSB usually comes with smaller sampling error, since it does not make asymptotic assumptions on $T$. 
However, the current approaches~\citep[see, e.g., methods in ][]{DeBortoli2021diffusion, Shi2023diffusion, Chen2022likelihood} for solving Equation~\eqref{equ:dsb} can be computationally more demanding than the commonly used denoising score matching, as their estimation methods usually require storing sample paths over time, potentially resulting in higher training cost. 

\section{Conditional sampling with joint samples}
\label{sec:cond-sampling-joint}
In the previous section we have seen the gist of generative sampling as well as two constructions of the generative samplers. 
However, they cannot be directly used to sample the conditional distribution $\pi(\cdot \cond y)$, since they need samples of $\pi(\cdot \cond y)$ to estimate Equations~\eqref{equ:bwd} and~\eqref{equ:dsb}. 
Recently, \citet{Vargas2023denoising} and~\citet{Phillips2024particle} have developed generative samplers without using samples of the target, but when applied to conditional sampling, we would have to re-train their models every time the condition $y$ changes. 
Hence, in this section we show how to apply the two constructions (i.e., Anderson and DSB) to train conditional samplers by sampling the \emph{joint distribution} $\pi_{X,Y}$ which is a reasonable assumption for many applications (e.g., image inpainting, super-resolution, and class-conditioning where we have paired data). 
We start by a heuristic idea, called Doob's bridging, and then we show how to extend it to a generic framework that covers many other conditional generative samplers.

\subsection{Doob's bridging}
\label{sec:doobs-bridging}
Recall that we aim to sample $\pi(\cdot \cond y)$, and assume that we can sample the joint $\pi_{X,Y}$. 
Denote by $q_{t \cond s}(\cdot \cond v)$ the conditional distribution of $U(t)$ conditioned on $U(s)=v$ for $t > s$. 
The idea is now to construct the reversal in such a way that 
\begin{equation}
	q_{T \cond 0}(x \cond y) = \pi(x \cond y)
	\label{equ:cond-rev-aim}
\end{equation}
for all $x\in\R^d$. 
If we in addition assume that the dimensions of $X$ and $Y$ match, then an immediate construction that satisfies the requirement above is 
\begin{equation}
	\diff U(t) = f(U(t), t) \diff t + g(t) \diff W(t), \quad \bigl( U(T), U(0) \bigr) \sim \pi_{X, Y},
	\label{equ:doob-bwd}
\end{equation}
where the terminal and initial jointly follow $\pi_{X, Y}$ by marginalising out the intermediate path. 
Thus, sampling $\pi(\cdot \cond y)$ simplifies to simulating the SDE in Equation~\eqref{equ:doob-bwd} with initial value $U(0) = y$. 
Directly finding such an SDE is hard, but we can leverage Anderson's construction to form it as the reversal of a forward equation
\begin{equation}
	\diff X(t) = a(X(t), Y, t) \diff t + b(t) \diff W(t), \quad \bigl( X(0), X(T) \bigr) \sim \pi_{X, Y}. 
	\label{equ:doob-fwd}
\end{equation}
Now we have a few handy methods to explicitly construct the forward equation above. 
One notable example is via Doob's $h$-transform~\citep{Williams2000Vol2} by choosing $X(0) = X$ and 
\begin{equation}
	a(x, Y, t) = \mu(x) + \Gamma(t)\grad_x\log h_T(Y, x, t), 
\end{equation}
where $\mu$ is the drift function of another (arbitrary) reference SDE
\begin{equation*}
	\diff Z(t) = \mu(Z(t)) \diff t + b(t) \diff W(t),
\end{equation*}
and the $h$-function $h_T(y, x, t) \coloneqq p_{Z(T) \cond Z(t)}(y \cond x)$ is the conditional density function of $Z(T)$ with condition at $t$. 
With this choice, it is implied that $X(T) = Y$ almost surely~\citep{Williams2000Vol2}, and thus the desired $\bigl( X(0), X(T) \bigr) \sim \pi_{X, Y}$ is satisfied. 
However, we have to remark a caveat that the process $t\mapsto X(t)$ in Equation~\eqref{equ:doob-fwd} \emph{marginally is not} a Markov process, since its SDE now invokes $Y$ which is correlated with the process across time. 
But on the other hand, this problem can be solved by an extension $\diff Y(t) = 0$, $Y(0)=Y$, such that $(X(t), Y(t))$ is jointly Markov. 
We can then indeed make use of Anderson's construction to learn the reversal in Equation~\eqref{equ:doob-bwd}. 
This approach was exploited by~\citet{Somnath2023} who coined the term ``aligned Schr\"{o}dinger bridge''. 
However, we note that the approach does not solve the DSB problem in Equation~\eqref{equ:dsb}:
In DSB we aim to find the optimal coupling under a path constraint, but in this approach, the coupling is already fixed by the given $\pi_{X, Y}$.

Generative conditional sampling with Doob's bridging is summarised as follows.

\begin{algorithm}[Doob's bridging conditional sampling]
	\label{alg:doob-pinning}
	The Doob's bridging conditional sampling consists in training and sampling a generative diffusion as follows. 
	\begin{enumerate}
		\item[Training.] Draw $\lbrace (X^{i}, Y^{i}) \rbrace_{i=1}^n \sim \pi_{X, Y}$, and then simulate paths of Equation~\eqref{equ:doob-fwd} with these samples as their respective initials. 
		Based on these paths, learn the reversal in Equation~\eqref{equ:doob-bwd} by, for instance, adapting the score matching in Equation~\eqref{equ:score-matching}. 
		Note that since the forward drift depends on $Y$, the reversal will also depend on $Y$.
		\item[Sampling.] After the reversal is estimated. 
		For any given condition $y$, simulate Equation~\eqref{equ:doob-bwd} with initial value $U(0)=y$. 
		Then it follows that $U(T) \sim \pi(\cdot \cond y)$. 
	\end{enumerate}
\end{algorithm}

From the algorithm above we see that the approach is as straightforward as standard generative sampling for unconditional distributions, with an additional involvement of $Y$ in the forward and reverse equations; It adds almost no computational costs. 
Moreover, the algorithm does not assume $T\to\infty$ unlike many other generative samplers. 
However, the algorithm suffers from a few non-trivial problems due to the use of Doob's transform to exactly pin two points. 
First, Doob's $h$-function is typically not available in closed form when the reference process is nonlinear.
Although it is still possible to simulate Equation~\eqref{equ:doob-fwd} without knowing $h_T$ explicitly~\citep{Schauer2017}, and to approximate $h_T$~\citep{Bagmark2022, Baker2024score}. This hinders the use of denoising score matching which is a computationally efficient estimator, as the conditional density $p_{t \cond s}$ of the forward equation is consequently intractable. 
%Although it is possible to simulate Equation~\eqref{equ:doob-fwd} even not knowing $h_T$ explcitly, 
Second, the function $h_T(\cdot, \cdot, T)$ at time $T$ is not defined, and moreover, the forward drift is rather stiff around that time.
This induces numerical errors and instabilities that are hard to eliminate. 
Finally, the approach assumes that the dimensions of $X$ and $Y$ are the same, which limits the range of applications. 
Recently,~\citet{Denker2024deft} generalise the method by exploiting the likelihood $\pi_{Y \cond X}$ in such a way that the two points do not have to be exactly pinned. 
This can potentially solve the problems mentioned above.

\subsection{Generalised joint bridging}
\label{sec:generalised-bridging}
Recall again that we aim to find a pair of forward and reversal SDEs such that the reversal satisfies Equation~\eqref{equ:cond-rev-aim}. 
In the previous section, Doob's transform was exploited for that  purpose, but the approach comes with several problems due to the use of Doob's $h$-transform for \emph{exactly} pinning the data random variables $X$ and $Y$. 
In this section, we present a generalised forward-reverse SDE that does not require exact pinning. 

The core of the idea is to introduce an auxiliary variable $V\in\R^{d_y}$ in the joint $\bigl( U(0), U(T), V \bigr)$, such that we can sample $q_{T, 0 \cond V}(\cdot \cond y)$, and that the marginal at $T$ satisfies $q_{T \cond V}(\cdot \cond y) = \pi(\cdot \cond y)$ (cf.  Equation~\eqref{equ:cond-rev-aim}), for instance, $V = Y$ almost surely.
More precisely, the reversal that we desire is
\begin{equation}
	\diff U(t) = f(U(t), V, t) \diff t + g(t) \diff W(t), \quad (U(0), V) \sim \refmeasure, \quad (U(T), V) \sim \pi_{X, Y},
	\label{equ:joint-bwd}
\end{equation}
starting at any reference $\refmeasure$, so that $\int q_{T \cond 0, V}(\cdot \cond u, y) \refmeasure(\diff u \cond y) = q_{T \cond V}(\cdot \cond y) = \pi(\cdot \cond y)$. 
With this reversal, if we can sample $U(0) \sim \pi_{\mathrm{ref}}(\cdot \cond y)$, then consequently sample $U(T) \sim \pi(\cdot \cond y)$ by simulating the SDE.
Define the forward equation that paired with the reversal by 
\begin{equation}
	\diff X(t) = a(X(t), Y, t) \diff t + b(t)\diff W(t), \quad (X(0), Y) \sim \pi_{X, Y}, \quad (X(T), Y) \sim \refmeasure, 
	\label{equ:joint-fwd}
\end{equation}
where the reference measure $\refmeasure = p_{T, Y}$ is the joint distribution of $(X(T), Y)$. 
Doob's bridging is a special case, in the way that we have coerced $X(T) = Y$ as the reference. 
Also note that we can extend Equations~\eqref{equ:joint-bwd} and~\eqref{equ:joint-fwd} to Markov processes with $\diff Y(t)=0$ in the same way as in Doob's bridging. 
To ease later discussions, let us first summarise the conditional sampling by the joint bridging method as follows. 

\begin{algorithm}[Joint bridging conditional sampling]
	\label{alg:joint-bridging}
	Let the forward and reversal in Equations~\eqref{equ:joint-fwd} and~\eqref{equ:joint-bwd} be given (i.e., training the generative sampler is complete). 
	Sample $U(0) \sim \refmeasure(\cdot \cond y)$ and simulate the reversal at $T$, then $U(T) \sim \pi(\cdot \cond y)$. 
\end{algorithm}

The remaining question is how to identify the generative sampler to use in Algorithm~\ref{alg:joint-bridging}. 
Now looking back at Equations~\eqref{equ:fwd} and~\eqref{equ:bwd}, we see that the algorithm is essentially an unconditional generative sampler that operates on the joint space of $X$ and $Y$. 
Hence, we can straightforwardly choose either Anderson's construction or DSB to obtain the required reversal in Equation~\eqref{equ:joint-bwd}. 
With Anderson's approach, the reversal's drift function becomes
\begin{equation}
	f(u, v, t) = -a(u, v, T - t) + \Gamma(T - t)\grad_u\log p_{T-t \cond Y}(u \cond v), 
	\label{equ:cond-rev-drift}
\end{equation}
where we see that the marginal score in Equation~\eqref{equ:rev-drift-anderson} now turns into a conditional one. 
This conditional score function is the major hurdle of the construction, as the conditional score is often intractable. 
\citet{Song2021scorebased} utilise the decomposition $\grad_u\log p_{t \cond Y}(u \cond v) = \grad_u \log p_{t}(u) + \grad_u\log p_{Y \cond t}(v \cond u)$, where the two terms on the right-hand side can be more accessible. 
Take image classification for example, where $X$ and $Y$ stand for the image and label, respectively. 
The generative score $\grad_u \log p_{t}(u)$ is learnt via standard score matching, while the likelihood score $\grad_u\log p_{Y \cond t}(v \cond u)$ can be estimated by training an image classifier with categorical $p_{Y \cond t}$ for all $t$. 
However, this is highly application dependent, and developing a generic and efficient estimator for the conditional score remains an active research question~\citep[see, e.g., approximations in][]{Chung2023diffusion, Song2023pseudoinverseguided}. 

Another problem of using Anderson's construction is again the common assumption $T\to\infty$, as we need to sample the conditional reference $p_{T \cond V}=\refmeasure(\cdot \cond y)$. 
For example, \citet{Luo2023IRSDE} choose $a(x, Y, t) = \theta \, (Y - x)$ for some $\theta>0$, implying that $\refmeasure(\cdot \cond y) \approx \mathrm{N}(\cdot \cond y, \sigma^2)$ is approximately a stationary Gaussian. 
Under this choice, the resulting forward equation is akin to that of Doob's bridging but instead pins between $X$ and a Gaussian-noised $Y$. 

As an alternative to Anderson's construction, we can employ DSB to search for a pair of Equations~\eqref{equ:joint-bwd} and~\eqref{equ:joint-fwd} that exactly bridges between $\pi_{X, Y}$ and a given $\refmeasure$, within a finite horizon $T$. 
Formally, we are looking for a process 
\begin{equation}
	\begin{split}
		\diff X(t) &= a(X(t), Y(t), t) \diff t + b(t) \diff W(t), \quad \diff Y(t) = 0, \\
		\bigl(X(0), Y(0)\bigr) &\sim \pi_{X, Y}, \quad  \bigl(X(T), Y(0)\bigr) \sim \refmeasure, 
		\label{equ:cond-dsb-fwd}
	\end{split}
\end{equation}
whose path measure $\bbP$ minimises $\kl{\cdot}{\Q}$, where $\Q$ is the measure of a reference process
\begin{equation}
	\begin{split}
		\diff Z(t) &= \mu(Z(t), R(t), t) \diff t + \sigma(t) \diff W(t), \quad \diff R(t) = 0, \\
		\bigl( Z(0), R(0) \bigr) &\sim \pi_{X, Y}.
	\end{split}  
\end{equation}
Due to the fact that $\refmeasure$ is now allowed to be arbitrary, we can choose a  $\refmeasure$ such that $\refmeasure(\cdot \cond y)$ is a reasonable approximation to $\pi(\cdot \cond y)$ that is easy to sample from (e.g., Laplace approximation). 

Currently there are two (asymptotic) numerical approaches to solve for the (conditional) DSB in the two equations above. 
The approach developed by~\cite{DeBortoli2021diffusion} aims to find a sequence $\bbP^{i}$ that converges to $\bbP$ as $i\to\infty$ via half-bridges
\begin{align}
	\bbP^{2 \, i - 1} &= \argmin_{\bbP_T = \refmeasure} \klbig{\bbP}{\bbP^{2 \, i - 2}}, \\
	\bbP^{2 \, i} &= \argmin_{\bbP_0 = \pi_{X, Y}} \klbig{\bbP}{\bbP^{2 \, i - 1}},
	\label{equ:csb-fixed-point}
\end{align}
starting at $\bbP^0 \coloneqq \Q$. 
The solution of each half-bridge can be approximated by conventional parameter estimation methods in SDEs. 
As an example, in Equation~\eqref{equ:csb-fixed-point}, suppose that $\bbP^{2 \, i-1}$ is obtained and that we can sample from it, then we can approximate $\bbP^{2 \, i}$ by parametrising the SDE in Equation~\eqref{equ:cond-dsb-fwd} and then estimate its parameters via drift matching between $\bbP^{2 \, i}$ and $\bbP^{2 \, i-1}$. 
However, a downside of this approach is that the solution does not exactly bridge $\pi_{X, Y}$ and $\refmeasure$ at any $i$ albeit the asymptoticity: 
it is either $\pi_{X, Y}$ or $\refmeasure$ that is preserved, not both. 
This problem is solved by~\citet{Shi2023diffusion}. 
They utilise the decomposition $\bbP = \pi^\star \, \Q_{\cond 0, T}$, where $\pi^\star$ solves the optimal transport $\min_{\nu}\lbrace \kl{\nu}{\Q_{0, T}} \colon \nu_0 = \pi_{X, Y}, \nu_T = \refmeasure \rbrace$, $\nu$ is a coupling of marginals $\nu_0$ and $\nu_T$, and $\Q_{\cond 0, T}$ stands for the measure of the reference process conditioning on its initial and terminal values at times $0$ and $T$ (e.g., Doob's $h$-transform). 
Denote $\bbP_{0, T}$ as the marginal of $\bbP$ at both $t=0$ and $t=T$, the resulting algorithm is then alternating between
\begin{equation}
	\begin{split}
		\bbP^{2 \, i - 1} &= \mathrm{proj}_{\mathcal{M}}\bigl(\bbP^{2 \, i - 2}\bigr), \\
		\bbP^{2 \, i} &= \bbP^{2 \, i - 1}_{0, T} \, \Q_{\cond 0, T}, 
	\end{split}
\end{equation}
starting at $\bbP^0 \coloneqq \widetilde{\pi} \, \Q_{\cond 0, T}$, where $\widetilde{\pi}$ is any coupling of $\pi_{X,Y}$ and $\refmeasure$, for instance, the trivial product measure $\pi_{X, Y} \times \refmeasure$. 
In the iteration above, $\mathrm{proj}_{\mathcal{M}}\bigl(\bbP^{2 \, i - 2}\bigr)$ approximates $\bbP^{2 \, i - 2}$ as a Markov process, since the previous $\bbP^{2 \, i - 2}$ is not necessarily Markovian. 
Moreover, \citet{Shi2023diffusion} choose a particular projection $\mathrm{proj}_{\mathcal{M}}\bigl(\bbP^{2 \, i - 2}\bigr) = \argmin_{\bbP\in\mathcal{M}} \klbig{\bbP^{2 \, i - 2}}{\bbP}$, such that their time-marginal distributions match, and therefore preserve the marginals $\bbP^i_0 = \pi_{X, Y}$ and $\bbP^i_T = \refmeasure$ at any $i$. 
In essence, this algorithm builds a sequence on couplings $\bbP^{i}_{0, T}$ that converges to $\pi^\star$, in conjunction with Doob's bridging. 
Although the method has to simulate Doob's $h$-transform $\Q_{\cond 0, T}$ which can be hard for complex reference processes, the cost is marginal in the context. 

Compared to the iteration in Equation~\eqref{equ:csb-fixed-point}, this projection-bridging method is particularly useful in the generative diffusion scenario, since we mainly care about the marginal distribution and not so much about the actual interpolation path. 
That said, at any iteration $i$, the solution, is already a valid bridge between $\pi_{X, Y}$ and $\refmeasure$ at disposal for generative sampling, even though it is not yet a solution to the Schr\"{o}dinger bridge. 

\subsection{Forward-backward path bridging}
\label{sec:fbs}
The previous joint bridging method (including Doob's bridging as a special case) aims to find a reversal such that the conditional terminal is $q_{T \cond V}(\cdot \cond y) = \pi(\cdot \cond y)$. 
In essence, it thinks of sampling the target as simulating a reverse SDE. 
In this section, we show another useful insight, by viewing sampling the target as simulating a stochastic filtering distribution~\citep{Corenflos2024conditioning, Dou2024, Trippe2023protein}. 
Formally, this approach sets base on 
\begin{equation}
	\begin{split}
		\diff X(t) &= a_1(X(t), Y(t), t) \diff t + b_1(t) \diff W_1(t), \\
		\diff Y(t) &= a_2(X(t), Y(t), t) \diff t + b_2(t) \diff W_2(t), \\
		\bigl(X(0), Y(0)\bigr) &\sim \pi_{X, Y}, 
	\end{split}
	\label{equ:fbs-fwd}
\end{equation}
and the core is the identity
\begin{equation}
	\begin{split}
		&\int p_{X(0) \cond Y(0), Y_{(0, T]}, X(T)}(\cdot \cond y, y_{(0, T]}, x_T) \, p_{X(T), Y_{(0, T]} \cond Y(0)}(\diff x_T, \diff y_{(0, T]} \cond y) \\
		&= p_{X(0) \cond Y(0)}(\cdot \cond y) = \pi(\cdot \cond y),
		\label{equ:fbs-identity}
	\end{split}
\end{equation}
where the integrand $p_{X(0) \cond Y(0), Y_{(0, T]}, X(T)}(\cdot \cond y, y_{(0, T]}, x_T)$ is the filtering distribution (in the reverse-time direction) with initial value $x_T$ and measurement $y_{[0, T]} \coloneqq [ y,  y_{(0, T]} ]$. 
The identity describes an ancestral sampling that, for any given condition $y$, we first sample $\bigl(X(T), Y_{(0, T]}\bigr) \sim p_{X(T), Y_{(0, T]} \cond Y(0)}(\cdot \cond y)$ and then use it to solve the filtering problem of the reversal, hence the name ``forward-backward path bridging''. 
However, since these two steps cannot be computed exactly, we show how to approximate them in practice in the following. 

The distribution $p_{X(T), Y_{(0, T]} \cond Y(0)}(\cdot \cond y)$ is in general not tractable. 
In~\citet{Dou2024} and~\citet{Trippe2023protein} they make assumptions facilitating the following approximation
\begin{equation}
	p_{X(T), Y_{(0, T]} \cond Y(0)} = p_{X(T) \cond Y_{[0, T]}} \, p_{Y_{(0, T]} \cond Y(0)} \approx p_{X(T) \cond Y(T)} \, p_{Y_{(0, T]} \cond Y(0)}. 
	\label{equ:trippe-dou}
\end{equation}
More specifically, they assume that 1) the SDE in Equation~\eqref{equ:fbs-fwd} is separable (i.e., $a_1$ does not depend on $Y$ while $a_2$ does not depend on $X$), so that they can independently simulate $p_{Y_{(0, T]} \cond Y(0)}(\cdot \cond y)$ leaving away the part that depends on $X$, and that 2) the SDE is stationary enough for large $T$ such that $p_{X(T) \cond Y_{[0, T]}} \approx p_{X(T) \cond Y(T)}$. 
Furthermore, the (continuous-time) filtering distribution $p_{X(0) \cond Y(0), Y_{(0, T]}, X(T)}$ is also intractable. 
In practice, we apply sequential Monte Carlo~\citep[SMC,][]{ChopinBook2020} to approximately sample from it. 
This induces error due to 3) using a finite number of particles. 
Since continuous-time particle filtering~\citep{Bain2009} is particularly hard in the generative context, we often have to 4) discretise the SDE beforehand, which again introduce errors, though simulating SDEs without discretisation is possible~\citep{Beskos2005Exact, Blanchet2020}. 
These four points constitute the main sources of errors of this type of method.

The work of~\citet{Corenflos2024conditioning} eliminates all the errors except for that of the SDE discretisation. 
The idea is based on extending the identity in Equation~\eqref{equ:fbs-identity} as an invariant Markov chain Monte Carlo (MCMC) kernel $\mathcal{K}_y$ by
\begin{align}
	&(\mathcal{K}_yh)(\cdot) \\
	&= \int p_{X(0) \cond Y(0), Y_{(0, T]}, X(T)}(\cdot \cond y, y_{(0, T]}, x_T) \int p_{X(T), Y_{(0, T]} \cond X(0), Y(0)}(\diff x_T, \diff y_{(0, T]} \cond z, y) \, h(\diff z), \nonumber
\end{align}
where we see that the kernel is indeed $\pi(\cdot \cond y)$-variant: $\mathcal{K}_y(p_{X(0) \cond Y(0)}) = p_{X(0) \cond Y(0)} = \pi_{X\cond Y}$. 
Moreover, we can replace the filtering distribution with another MCMC kernel by, for instance, gradient Metropolis--Hasting~\citep{Titsias2018}, or more suitable in this context, conditional sequential Monte Carlo~\citep[CSMC,][]{Andrieu2010particle}. 
The algorithm finally works as a Metropolis-within-Gibbs sampler as follows. 

\begin{algorithm}[Gibbs filtering of conditional generative diffusion]
	\label{alg:gibbs-zz-ac}
	Let $X^0 \sim \pi(\cdot \cond y)$, and $q$ be the law of the reversal of Equation~\eqref{equ:fbs-fwd}. 
	For $i=1,2,\ldots$, do 
	\begin{enumerate}
		\item \label{alg:gibbs-zz-ac-fwd} simulate forward $\bigl( X_T^i, Y^i_{(0, T]} \bigr) \sim p_{X(T), Y_{(0, T]} \cond Y(0), X(0)}(\cdot \cond y, X^{i-1})$, 
		\item set the reverse path $V_{[0, T]}^i = Y_{[T, 0]}^i$, 
		\item \label{alg:gibbs-zz-ac-bwd} simulate $X^i \sim q_{U(T) \cond V_{[0, T]}, U(0)}(\cdot \cond V^i_{[0, T]}, X_T^i)$ using CSMC.
	\end{enumerate}
	Then $X^i \sim \pi(\cdot \cond y)$ for any $i$. 
\end{algorithm}

The Gibbs filtering-based conditional sampler above has significantly less errors compared to~\citet{Trippe2023protein} and~\citet{Dou2024} as empirically shown in~\citet{Corenflos2024conditioning}. 
In a sense, the Gibbs sampler has transformed the two major errors due to $T\to\infty$ and using a finite number of particles, to the increased autocorrelation in Algorithm~\ref{alg:gibbs-zz-ac}. 
When $T$ is small, the samples are more correlated, but still statistically exact nonetheless. 
Moreover, the sampler does not assume the separability of the diffusion model, meaning that in the algorithm we are free to use 
any kind of reversal construction (e.g., Anderson or DSB). 
On the other hand, \citet{Trippe2023protein} and~\citet{Dou2024}'s approach does not instantly apply on DSBs which are rarely separable. 
The cost we pay for using the Gibbs sampler is mainly the implementation and statistical efficiency of the filtering MCMC in~\ref{alg:gibbs-zz-ac-bwd} of Algorithm~\ref{alg:gibbs-zz-ac}.
It is also an open question for simulating the filtering MCMC in continuous-time without discretising the SDE. 

\paragraph{Comparing the filtering-based approaches to the joint bridging in Section~\ref{sec:generalised-bridging}.} These two methods mainly differ in the utilisation of their reversals, and they shine in different applications. 
More precisely, the joint bridging method aims to \emph{specifically construct} a pair of forward and backward models that serves the conditional sampling, whereas the filtering-based methods work on any \emph{given} forward-backward pair. 
This grants the filtering methods an upper hand for some applications as training-free samplers. 
Take image inpainting for example, where $X$ and $Y$ stand for the observed and to-paint parts, respectively. 
The joint bridging method has to specifically train for a pair of forward and backward models for this inpainting problem. 
On the other hand, the filtering-based methods can take on any pre-trained unconditional diffusion models for the images (see Section~\ref{sec:generative-sampling}), and plug-and-play without any training~\citep[see, e.g.,][Sec. 4.3]{Corenflos2024conditioning}. 
This is because in the inpainting application $X$ and $Y$ completely factorises the unconditional diffusion model, and~\citet{Dou2024} and~\citet{Cardoso2024monte} recently show that it can further generalise to linear inverse problems. 

However, for applications when the training-free feature cannot be utilised, the joint bridging method is easier to implement and usually runs with less computational costs. 
The joint bridging method does not keep a path of $Y_{[0, Y]}$, and thus it does not need to solve a (reverse) filtering problem which is significantly harder than simulating a reverse SDE. 
Furthermore, the joint bridging method immediately allows for discrete condition $Y$, while this is yet an interesting open question for the filtering-based methods. 
The discrete extension of generative diffusions in~\citet{Campbell2022, Benton2024} may provide insights to enable the discrete extension. 

\section{Conditional sampling with explicit likelihood}
\label{sec:cond-sampling-likelihood}
In Section~\ref{sec:cond-sampling-joint} we have investigated a class of methods that leverage the information from the joint distribution $\pi_{X, Y}$ to build up conditional samplers for $\pi(\cdot \cond y)$. 
However, for some applications we may only be able to sample the marginal $\pi_X$. 
For instance, in image classification we usually have vast amount of images at hand, but pairing them with labels requires extensively more efforts.
Thus, in this section we show how to construct conditional samplers \emph{within prior generative diffusions}, based on exploiting the additional information from the likelihood $\pi(y \cond \cdot)$. 
%In particular, we focus the methods based on Feynman--Kac models with sequential Monte Carlo. 

\subsection{Approximate conditional sampling with conditional score}
Recall the conditional score $\grad_x \log p_{t \cond Y}(x \cond y)$ in Equation~\eqref{equ:cond-rev-drift}, and that we can decompose it into $\grad_x\log p_{t \cond Y}(x \cond y) = \grad_x \log p_{t}(x) + \grad_x\log p_{Y \cond t}(y \cond x)$. 
The second term, the measurement score $\grad_x\log p_{Y \cond t}(y \cond x)$ is closely related to the target likelihood in the way that $p_{Y \cond t}(y \cond x) = \int \pi(y \cond x_0) \, p_{0 \cond t}(x_0 \cond x) \diff x_0$.
This implies that we can approximate the measurement score by exploiting the likelihood, and therefore perform conditional sampling by simulating the SDE associated with the measurement score as per Algorithm~\ref{alg:joint-bridging}. 
The most celebrated approach is perhaps the diffusion posterior sampling~\citep{Chung2023diffusion}.
It approximates the integral $\int \pi(y \cond x_0) \, p_{0 \cond t}(x_0 \cond x) \diff x_0 \approx \pi(y \cond m_0(x))$ by a first-order linearisation with mean $m_0(x)=\expec{X(0) \cond X(t)=x}$. 
This approach works asymptotically accurate as $t\to 0$, but large errors may accumulate in the early steps. 
Recently there has been a lot of work presented on how to  effectively approximate the measurement score depending on the intended application. 
We refer the readers to, for instance, \citet{Daras2024survey} for a detailed review of existing approximations. 

The major challenge of this approach is that we can barely compute the measure score exactly for most models, and thus it will introduce statistical errors to the conditional samplers. 
In the next section, we show how to resolve this statistical bias by wrapping these approximate measurement scores as effective proposals in a Feynman--Kac model and sequential Monte Carlo sampling. 

\subsection{Conditional generative Feynman--Kac models}
\label{sec:twisted}
Let us assume that we are given a pair of forward-backward diffusion models (using arbitrary reversal construction) that targets $\pi_X$ as in Equations~\eqref{equ:fwd} and~\eqref{equ:bwd}. 
For simplicity, let us change to work with the models at discrete times $k=0,1, 2,\ldots, N$ via
\begin{equation*}
	p_{0:N}(x_{0:N}) = p_0(x_0)\prod_{k=1}^N p_{k \cond k-1}(x_k \cond x_{k-1}), \quad q_{0:N}(u_{0:N}) = q_0(u_0)\prod_{k=1}^N q_{k \cond k-1}(u_k \cond u_{k-1}),
\end{equation*}
where $p_0(\cdot) = \pi_X(\cdot) = q_{N}(\cdot)$, and $q_0(\cdot) = \pi_{\mathrm{ref}}(\cdot) = p_N(\cdot)$. 
The distribution $p_{k\cond k-1}$ stands for the forward transition between discrete times $k-1$ and $k$.
Since we can sample $\pi_X$ and evaluate the likelihood $\pi(y \cond \cdot)$, it is natural to think of using importance sampling to sample $\pi(\cdot \cond y)$. 
That is, if $\lbrace X^j \rbrace_{j=1}^J \sim \pi_X$ then $\lbrace (w_j, X^j) \rbrace_{j=1}^J \sim \pi(\cdot \cond y)$ with weight $w_j = \pi(y \cond X^j)$. 
However, the weights will barely be informative, as the prior samples are very unlikely samples from the posterior distribution in generative applications. 
Hence, in practice we generalise importance sampling within Feynman--Kac models~\citep{ChopinBook2020} allowing us to effectively sample the target with sequential Monte Carlo (SMC) samplers. 
Formally, we define the generative Feynman--Kac model by 
\begin{equation}
	\begin{split}
		Q_{0:N}(u_{0:N}) &\coloneqq \frac{1}{\ell_N} \, M_0(u_0) \, G_0(u_0) \prod_{k=1}^N M_{k \cond k-1}(u_{k} \cond u_{k-1}) \, G_k(u_k, u_{k-1}), \\
		\text{s.t.} \quad Q_N &\coloneqq \int Q_{0:N}(u_{0:N}) \diff u_{0:N-1} = \pi(\cdot \cond y), \quad M_0 = q_0 = \refmeasure, 
	\end{split}
	\label{equ:feynman-kac}
\end{equation}
where $\lbrace M_{k\cond k-1} \rbrace_{k=0}^N$ and $\lbrace G_k \rbrace_{k=0}^N$ are Markov kernels and potential functions, respectively, and $\ell_N$ is the marginal likelihood. 
Instead of applying naive importance sampling, we apply SMC to simulate the model, crucially, the target $Q_N$, as in the following algorithm. 

\begin{algorithm}[Generative SMC sampler]
	\label{alg:feynman-kac-smc}
	The sequential Monte Carlo samples the Feynman--Kac model in Equation~\eqref{equ:feynman-kac} as follows. 
	First, draw $U^1_0, U^2_0, \ldots, U^J_0 \sim M_0$ and set $w_0^j = G_0(U_0^j) \, / \, \sum_{i=1}^J G_0(U_0^i)$ for $j=1,2,\ldots,J$. 
	Then, for $k=1, 2, \ldots, N$ do
	\begin{enumerate}
		\item Resample $\bigl\lbrace (w_{k-1}^j, U_{k-1}^j) \bigr\rbrace_{j=1}^J$ if needed. 
		\item Draw $U_k^j \sim M_{k \cond k-1}(\cdot \cond U_{k-1}^j)$ and compute $\overline{w}_k^j = w_{k-1}^j\, G_k(U_k^j, U_{k-1}^j)$ for $j=1,2,\ldots, J$.
		\item Normalise $w_k^j = \overline{w}^j_k \, / \, \sum_{i=1}^J \overline{w}_k^i$ for $j=1,2,\ldots, J$.
	\end{enumerate}
	At the final step $N$, the tuples $\bigl\lbrace (w_N^j, U_{N}^j) \bigr\rbrace_{j=1}^J$ are weighted samples of $\pi(\cdot \cond y)$.
\end{algorithm}

Given only the marginal constraints (e.g., at $Q_N$ and $Q_0$), the generative Feynman--Kac model is not unique. 
A trivial example is by letting $M_0 = q_0$, $M_{k \cond k-1} = q_{k \cond k-1}$, $G_0 = G_1 = \cdots = G_{N-1} \equiv 1$, and $G_N(u_k, u_{k-1}) = \pi(y \cond u_k)$, where Algorithm~\ref{alg:feynman-kac-smc} recovers naive importance sampling which is not useful in reality. 
The purpose is thus to design $M$ and $G$ so that $Q_k$ continuously anneals to $\pi(\cdot \cond y)$ for $k=0,1,\ldots,N$ while keeping the importance samples in Algorithm~\ref{alg:feynman-kac-smc} effective. 
\citet{Wu2023practical}~and~\citet{Janati2024divide} show a particularly useful system of Markov kernels and potentials by setting
\begin{equation}
	M_0 = q_0, \quad G_0 = l_0, \quad M_{k \cond k-1}(u_k \cond u_{k-1}) \, G_k(u_k, u_{k-1}) = \frac{l_k(y, u_k) \, q_{k \cond k-1}(u_k \cond u_{k-1})}{l_{k-1}(y, u_{k-1})},
	\label{equ:feynman-kac-choice}
\end{equation}
where $l_k(y, u_k) \coloneqq \int \pi(y \cond u_N) \, q_{N \cond k}(u_N \cond u_k)\diff u_N$, and define $l_N(y, u_N) \coloneqq \pi(y \cond u_N)$ that reduces to the target's likelihood. 
Indeed, if we substitute Equation~\eqref{equ:feynman-kac-choice} back into Equation~\eqref{equ:feynman-kac}, the marginal constraint $Q_N$ is satisfied. 
This choice has at least two advantages that makes it valuable in the generative diffusion context. 
First, this construction is locally optimal, in the sense that if we also fix  $G_k(u_k, u_{k-1}) \equiv 1$ then $M_{k \cond k-1}(u_k \cond u_{k-1}) = q_{k \cond k-1, Y}(u_k \cond u_{k-1}, y) \propto l_k(y, u_k) \, q_{k \cond k-1}(u_k \cond u_{k-1})$ which is exactly the Markov transition of the conditional reversal in Equation~\eqref{equ:cond-rev-drift}. 
Consequently, all samples in Algorithm~\ref{alg:feynman-kac-smc} are equally weighted, and hence the SMC sampler is perfect. 
% explain why local
However, simulating the optimal Markov kernel is hard, as $l_k(y, u_k)$ is intractable (except when $k=N$). 
This gives rise to another advantage: even if we replace the exact $l_k$ by any approximation $\widehat{l}_k$ for $k=0,1,\ldots, N-1$, the Feynman--Kac model remains valid. 
Namely, the weighted samples of Algorithm~\ref{alg:feynman-kac-smc} converge to $\pi(\cdot \cond y)$ in distribution as $J\to\infty$ even if $\lbrace l_k \rbrace_{k=0}^{N-1}$ is approximate. 
However, the quality of the approximation $\widehat{l}_k$ significantly impacts the effective sample size of the algorithm, and recently there have been numerous approximations proposed.

Following Equation~\eqref{equ:feynman-kac-choice}, \citet{Wu2023practical} choose to approximate $l_k$ by linearisation, and let the Markov kernel be that of a Langevin dynamic that approximately leaves $q_{k\cond k-1, Y}$ invariant. 
More specifically, they choose
\begin{equation}
	\begin{split}
		M_{k \cond k-1}(u_k \cond u_{k-1}) &= \widehat{q}_k(u_k \cond u_{k-1}, y), \quad G_k(u_k, u_{k-1}) = \frac{\widehat{l}_k(y, u_k) \, q_{k \cond k-1}(u_k \cond u_{k-1})}{\widehat{l}_{k-1}(y, u_{k-1}) \, \widehat{q}_k(u_k \cond u_{k-1}, y)}, \\
		\widehat{l}_k(y, u_k) &\coloneqq \pi\bigl(y \cond m_N(u_k)\bigr), \\ 
		\widehat{q}_k(u_k \cond u_{k-1}, y) &\coloneqq \mathrm{N}\Bigl( u_k \cond u_{k-1} +  \delta_k \grad_{u_k}\log \bigl( \widehat{l}_k(y, u_k) \, q_{k \cond k-1}(u_k \cond u_{k-1}) \bigr), \sqrt{2} \, \delta_k \Bigr), 
	\end{split}
	\label{equ:feynman-kac-wu-trippe}
\end{equation}
where $m_N(u_k)$ stands for any approximate sample from $q_{N \cond k}(\cdot \cond u_k)$, and $\delta_k$ is the Langevin step size. 
Originally, \citet{Wu2023practical} use $m_N(u_k)$ as a sample of the reversal at $N$ starting at $u_k$, which often experiences high variance, but it can be improved by using $m_N(u_k)$ as the conditional mean of $q_{N \cond k}(\cdot \cond u_k)$ via Tweedie's formula~\citep{Song2023pseudoinverseguided}. 
See also~\citet{Cardoso2024monte} for an alternative construction of the proposal. 
While \citet{Wu2023practical}'s construction of the Feynman--Kac model is empirically successful in a number of applications, its main problem lies in its demanding computation. 
In particular, the proposal asks to differentiate through $\widehat{l}_k$ usually containing a denoising neural network. 

\citet{Janati2024divide} follow up the construction in Equation~\eqref{equ:feynman-kac-choice} and propose an efficient sampler based on partitioning the Feynman--Kac model into several contiguous sub-models, where each sub-model targets its next. 
As an example, when we use two blocks, $Q_{0:N}$ is divided into $Q^{(1)}_{0:j}$ and $Q^{(2)}_{j:N}$ for some $0\leq j\leq N$, where $Q^{(1)}_{0:j}$ starts at $q_0$ and targets $Q_j$, whereas $Q^{(2)}_{j:N}$ starts at $Q_{j}$ and targets $q_N$. 
Although the division looks trivial at a first glance, the catch is that within each sub-model we can form more efficient approximations to the Feynman--Kac Markov transition. 
More precisely, recall that in Equation~\eqref{equ:feynman-kac-wu-trippe}, $\widehat{l}_k(y, u_k)$ aims to approximate $\int \pi(y \cond u_N) \, q_{N \cond k}(u_N \cond u_k)\diff u_N$, the accuracy of which depends on the distance $N - k$. 
Now inside the sub-model $Q^{(1)}_{0:j}$, the terminal is $j$ instead of $N$, where $j - k$ is mostly smaller than $N - k$ for $0\leq k\leq j$. 
In addition, instead of running the SMC in Algorithm~\ref{alg:feynman-kac-smc}, they also propose a variational approximation to the Markov transition of the Feynman--Kac model in order to directly sample the model. 
Note that although \citet{Janati2024divide} set roots on linear Gaussian likelihood $\pi(y \cond \cdot)$, their method can straightforwardly generalise to other likelihoods as well, with suitable Gaussian quadratures applied.

The construction in~\citet{Cardoso2024monte} works efficiently when the target likelihood is Gaussian $\pi(y \cond x) = \mathrm{N}(y; H \, x, \sigma^2)$ for some operator $H\in\R^{d\times d_y}$, variance $\sigma^2$, and assuming $d>d_y$. 
Their core idea is to transform the conditional sampling problem into a noiseless inpainting problem for which they can come up with an efficient approximation to $l_k$. 
More precisely, they apply Algorithm~\ref{alg:feynman-kac-smc} with an inpainting likelihood $\pi(y\cond x) = \delta_{y}(\bar{x})$, where $\delta_{y}(\bar{x})$ is a Dirac delta, and $\bar{x}$ extracts the first $d_y$ dimensions of $x$. 
Their design of $l_k$ is empirically efficient for this inpainting likelihood.
To generalise for linear Gaussian likelihoods, they utilise a singular value decomposition $H = A \, \Sigma \, B^\trans$ and then equivalently work on a noisy inpainting likelihood $\mathrm{N}(A^\trans \, y; \Sigma\,\bar{z}, I)$, where $z\coloneqq B^\trans x$. 
They show that the noisy inpainting problem can be solved with an extension of the noiseless inpainting problem $\delta_{A^\trans \, y}(\Sigma \bar{z})$. 

A computationally simpler bootstrap construction of the generative Feynman--Kac model is
\begin{equation}
	\begin{split}
		M_{k \cond k-1}(u_k \cond u_{k-1}) &= q_{k \cond k-1}(u_k \cond u_{k-1}), \quad G_k(u_k, u_{k-1}) = \frac{\widehat{l}_{k}(y, u_{k})}{\widehat{l}_{k-1}(y, u_{k-1})}, \\
		\widehat{l}_k(y, u_k) &\coloneqq \pi(\lambda_k \, y \cond u_k),
		\label{equ:fk-simple}
	\end{split}
\end{equation}
where we simplify the Markov kernel by that of the unconditional reversal which is easier compared to $\widehat{q}_k(u_k \cond u_{k-1}, y)$ in Equation~\eqref{equ:feynman-kac-wu-trippe}. 
Additionally, the approximate $\widehat{l}_k$ is also simplified as the target likelihood evaluated at $u_k$ and a suitably scaled $\lambda_k \, y$ with $\lambda_N=1$ justified by the heuristics in~\citet[][Sec. 4.1]{Janati2024divide}. 
However, since the Markov proposal is farther from the optimal one, the resulting algorithm is statistically less efficient. 
When the problem of interests is not difficult, this bootstrap model may indeed be useful in practice (see our example in Section~\ref{sec:experiment}). 

\paragraph{Comparing the Feynman--Kac-based approaches to the joint bridging in Section~3\ref{sec:generalised-bridging}.}
As we have already pointed out, the Feynman--Kac model using Equation~\eqref{equ:feynman-kac-choice} indeed recovers Anderson's conditional SDE in Equation~\eqref{equ:cond-rev-drift}. 
The difference is that the Feynman--Kac model describes the evolution of the probability distribution (using an ensemble of weighted samples) between the reference and target, while the conditional SDE is formulated on the individual path. 
Essentially, they resemble the Lagrangian and Eulerian specifications, respectively, of a flow field~\citep{Santambrogio2015}. 
This gives the Feynman--Kac formalism an advantage in the way that it does not need to precisely compute the conditional score in Equation~\eqref{equ:cond-rev-drift} which is the main blocker of the conditional SDE approach. 
In other words, Algorithm~\ref{alg:feynman-kac-smc} transforms the errors in approximating the conditional score, to the effective sample size of the weighted samples. 
However, consequently the Feynman--Kac model pays additional computational costs, as it needs to store and process an ensemble of samples to describe the distribution. 
Another advantage of the Feynman--Kac model is that it does not need to train a dedicated conditional model, leveraging an analytical likelihood and a pre-trained reversal of $\pi_X$. 
Although the filtering-based method in Section~\ref{sec:fbs} can also work on pre-trained generative models, it's range of applications is more limited than Feynman--Kac.
It would be an interesting future question to ask whether the method can extend to unknown likelihoods, for instance, by applying the methods in~\citet{Sisson2007, Beaumont2009, Papamakarios2019,  Middleton2019}. 

\section{Pedagogical example}
\label{sec:experiment}

\begin{figure}[t!]
	\centering
	\includegraphics[width=.95\linewidth]{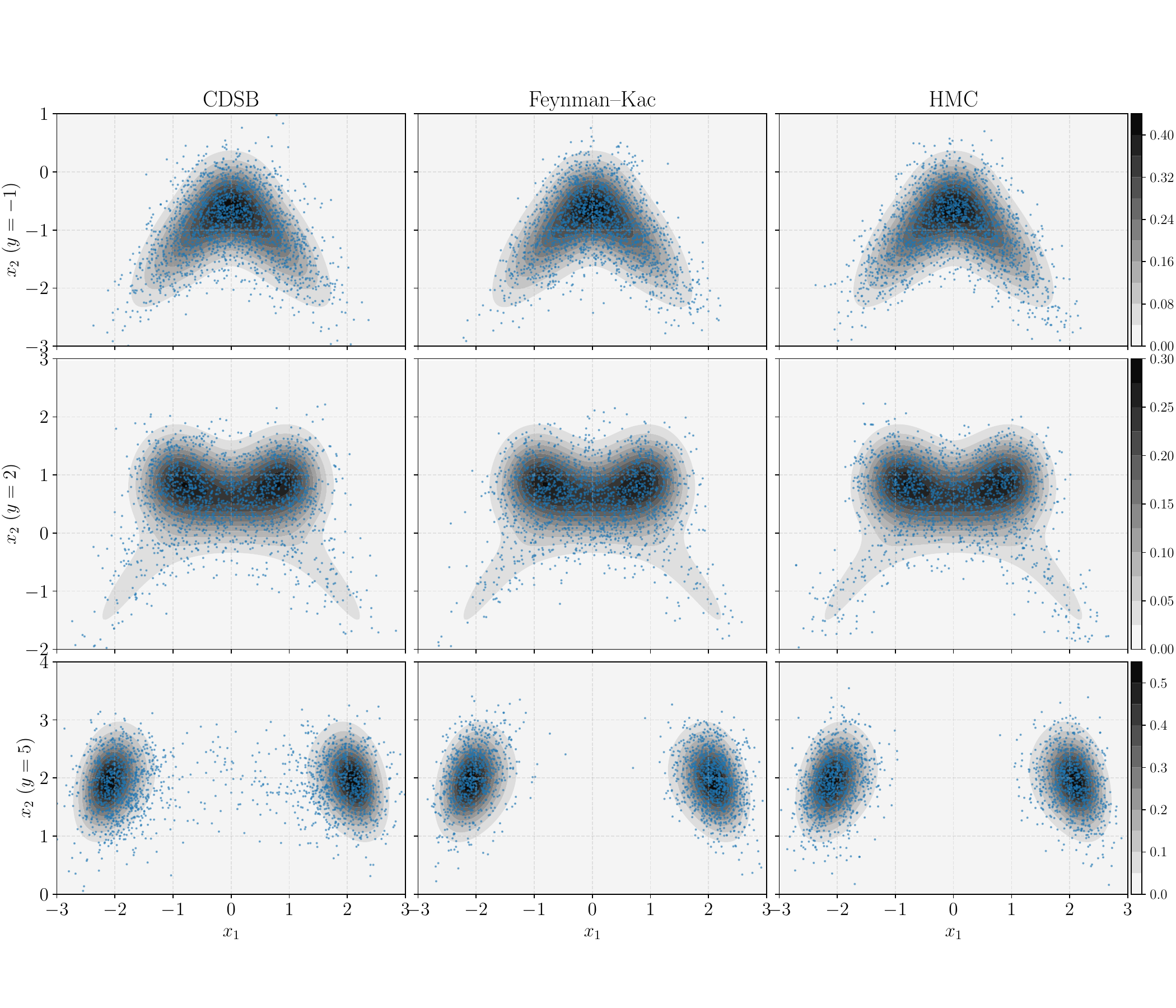}
	\caption{Illustration of conditional samples (in blue scatters) with three conditions on $y$, where the contour plots the true conditional density function. 
	Note that for each method we draw 10,000 samples, but we downsample it to 2,000 for visibility. 
	We see that all the three methods recover the true conditional distribution but with small biases. 
	For instance, CDSB with $y=5$ has biased samples between the two modes. }
	\label{fig:crescent}
\end{figure}
\begin{figure}[t!]
	\centering
	\includegraphics[width=.95\linewidth]{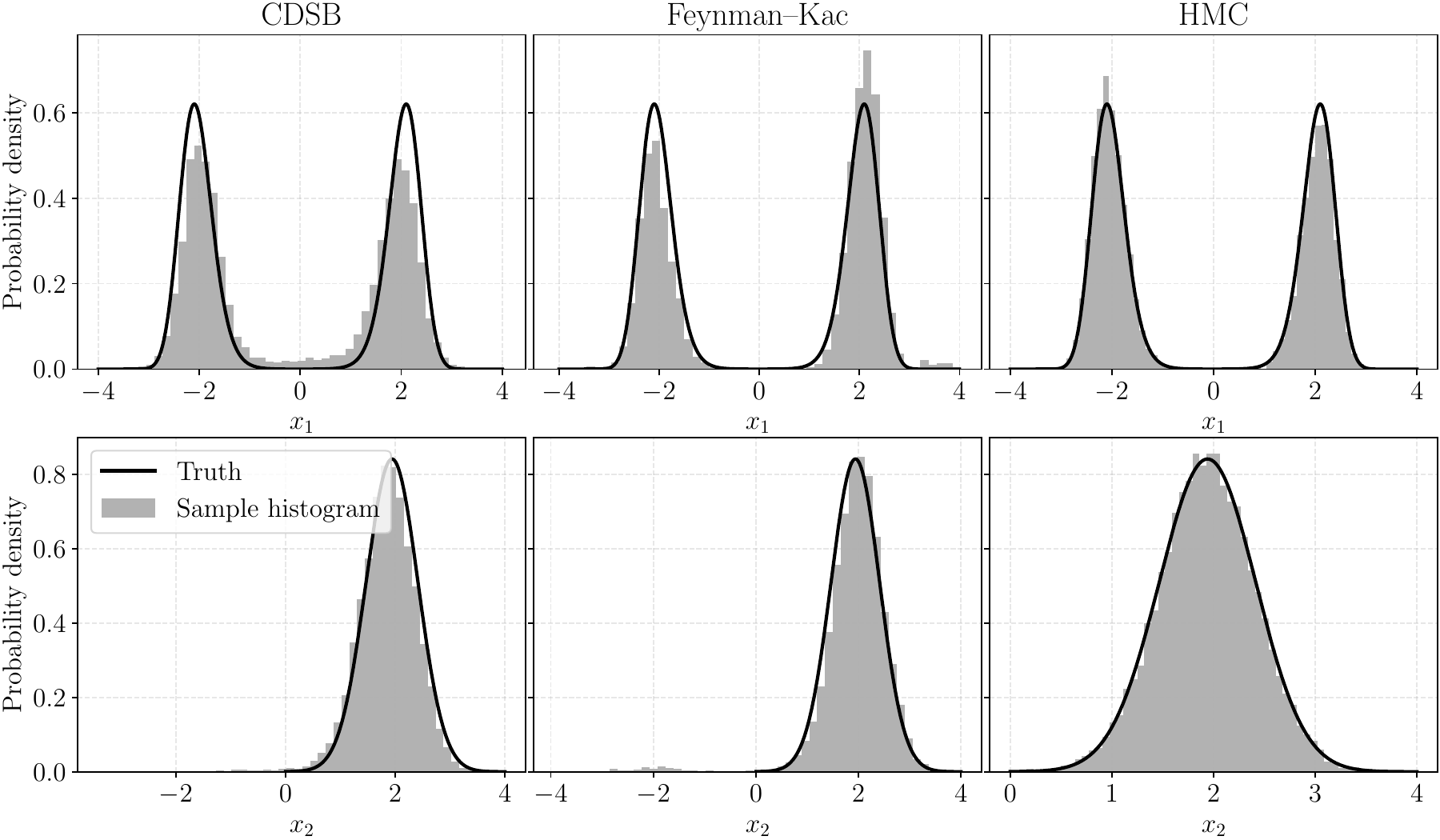}
	\caption{The marginal histograms with condition $y=5$ (corresponding to the last row in Figure~\ref{fig:crescent}). 
	We see that although the CDSB and Feynman--Kac methods recover the shape of the distribution, they are biased (e.g., note CDSB at around $x_1 = 0$ and Feynman--Kac at around $x_2 = -2$). }
	\label{fig:crescent-hist}
\end{figure}

In this section we present an example to illustrate 1) the joint bridging method with the Schr\"{o}dinger bridge construction in Equation~\eqref{equ:cond-dsb-fwd} and 2) the Feynman--Kac method using the construction in Equation~\eqref{equ:fk-simple}. 
%We focus on teaching how to use these two methods not comparison.
Importantly, we release our implementations, written with pedagogical intent.\footnote{See the code at \url{https://github.com/spdes/gdcs} implemented in JAX.}

The conditional sampling task we test here is a two-dimensional distribution given as follows. 
Let $\pi(x) = 0.5 \, \mathrm{N}(x\cond 0, v_0) + 0.5 \, \mathrm{N}(x\cond 0, v_1)$ be a two-dimensional Gaussian mixture, with covariances $v_0 = \begin{bmatrix}
	1 & 0.8 \\
	0.8 & 1
\end{bmatrix}$ and 
$v_1 = \begin{bmatrix}
	1 & -0.8 \\
	-0.8 & 1
\end{bmatrix}$, and define the likelihood as $\pi(y \cond x) \sim \mathrm{N}\bigl(y \cond x_2 + 0.5 \, (x_1^2 + 1), 0.5 \bigr)$. 
We brute-force compute the posterior distribution $\pi(\cdot \cond y)$ via trapezoidal integration for any given $y$ to compare the conditional samplers. 

To apply the joint bridging method with the Schr\"{o}dinger bridge construction (which we refer to as CDSB), the first step is to identify the forward Equation~\eqref{equ:cond-dsb-fwd} and its reversal, which constitutes the training part of the algorithm.
We choose a Brownian motion as the reference process, and apply the numerical method in Equation~\eqref{equ:csb-fixed-point} to solve for the forward and reversal with 15 iterations. 
This is achieved by parametrising the drifts of the forward and reversal, and then estimate their parameters by a drift matching formula~\citep[see, e.g.,][Alg. 1]{Shi2022Cond}.
The parametrisation uses a neural network with three layers of multilayer perceptions and sinusoidal embedding.
After training, the conditional sampling amounts to running Algorithm~\ref{alg:joint-bridging}. 
As for the Feynman--Kac method using the heuristic Equation~\eqref{equ:fk-simple}, we choose $\lambda_k \equiv 1$, stratified resampling, and train a DSB for $\pi_X$ with settings identical to those in CDSB.
The same neural network is used for both methods with time span $T=1$, $N=1\,000$ Euler discretisation steps, and reference distribution $\refmeasure = \mathrm{N}(0, 1)$. 
We also use Hamiltonian Monte Carlo (HMC) as a baseline to sample the posterior distribution with step size 0.35, unit mass, and 100 leap-frog integrations. 
Finally, we draw 10,000 Monte Carlo samples for the test. 

The results are shown in Figures~\ref{fig:crescent} and~\ref{fig:crescent-hist} tested with conditions $y=-1$, $2$, and $5$ leading to challenging distribution shapes. 
We see that in general both the CDSB and Feynman--Kac methods recover the true distribution to good extents, and that they are comparable to the MCMC method HMC. 
However, we note that the two generative methods are biased, particularly evidenced from their histograms. 
This makes sense, as the training for generative models can hardly be done exactly. 
Moreover, we also see that CDSB with $y=5$ is particularly erroneous between the two modes, as the training of CDSB uses the joint samples of $\pi_{X, Y}$, while the likelihood of $y=5$ is relatively low. 
This reflects that CDSB (or any other joint bridging method) may fail under extreme conditions. 
On the other hand, the Feynman--Kac method with SMC may also fail under extreme conditions, but it can be improved by using more particles (perhaps inefficiently).

\section{Conclusion}
\label{sec:conclusion}
In this article we have reviewed a class of conditional sampling schemes built upon generative diffusion models. 
We started by two foundational constructions for unconditional diffusion models: Anderson and Schr\"{o}dinger bridge. 
From there we explored two practical scenarios (i.e., whether the joint $\pi_{X, Y}$, or $\pi_X$ and $\pi_{Y \cond X}$ of the data distribution is accessible) to formulate the generative conditional samplers. 
In summary we have described three methodologies in detail. 
The first one, called joint bridging, involves training a dedicated generative diffusion that directly targets the conditional distribution $\pi(\cdot \cond y)$, framing the problem of conditional sampling as simulating a generative reverse diffusion. 
The second one, including diffusion Gibbs, also employs a generative diffusion but targets the joint $\pi_{X, Y}$, and it casts conditional sampling as solving a stochastic filtering problem. 
The last one, leveraging the likelihood $\pi_{Y\cond X}$ and a (pre-trained) generative diffusion that targets $\pi_X$, represents the conditional sampling as simulating a Feynman--Kac model. 
We have demonstrated the connections among these methodologies and highlighted their respective strengths in various applications. 
In the final section, we provided a pedagogical example illustrating the joint bridging and Feynman–Kac methods.

Finally, we close this paper with a few remarks and ideas for future work.
1) Akin to MCMC, the generative diffusion samplers would also need convergence diagnosis. 
The difference is that for MCMC we need to check if the sampler converges to the stationary phase, while for generative diffusions we are instead interested in measuring the biases. 
For instance, in our experiment the training of the generative samplers are empirically assessed by an expert, while MCMC are backed by ergodic theory.
Currently, to the best of our knowledge, it is not clear how to gauge if the trained generative sampler can be trusted in a principled way, in particular when it comes to high-dimensional problems. 
2) As illustrated in our experiment, sampling with extreme conditions can be problematic for generative diffusions. 
Addressing outliers is an important area for improving generative models.
3) Apart from the base methodologies of generative diffusions, the deep learning techniques~\citep[e.g., U-net, time embedding, see][for a longer list]{Song2020} are in our opinion the core that made them successful. 
Improving the structure and training of the involved neural networks are essential for making generative diffusion models functional in complex tasks, and we believe that this will remain a central focus in future developments.

\vskip6pt  % this margin comes from the template

\ack{This work was partially supported by the Kjell och M\"{a}rta Beijer Foundation, the Wallenberg AI, Autonomous Systems and Software Program (WASP) funded by the Knut and Alice Wallenberg Foundation and by the project \emph{Deep probabilistic regression -- new models and learning algorithms} (contract number: 2021-04301), by the Swedish Research Council.
}

\bibliographystyle{plainnat}
\bibliography{refs}

\end{document}